\documentclass[letterpaper]{article} 
\usepackage{arxiv}  
\usepackage{times}  
\usepackage{helvet}  
\usepackage{courier}  
\usepackage[hyphens]{url}  
\usepackage{graphicx} 
\usepackage{natbib}  
\usepackage{caption} 
\usepackage{algorithm}
\usepackage{algorithmic}
\usepackage{multirow}
\usepackage{amsmath}
\usepackage{amssymb}
\usepackage{newfloat}
\usepackage{listings}
\DeclareCaptionStyle{ruled}{labelfont=normalfont,labelsep=colon,strut=off} 
\floatstyle{ruled}
\newfloat{listing}{tb}{lst}{}
\floatname{listing}{Listing}
\title{FP-AbDiff: Improving Score-based Antibody Design by Capturing Nonequilibrium Dynamics through the Underlying Fokker–Planck Equation}
\author{
    Jiameng Chen\textsuperscript{\rm 1},
    Yida Xiong\textsuperscript{\rm 1},
    Kun Li\textsuperscript{\rm 1},
    Hongzhi Zhang\textsuperscript{\rm 1},
    Xiantao Cai\textsuperscript{\rm 1}\thanks{Corresponding authors.},
    Wenbin Hu\textsuperscript{\rm 3, 1}\footnotemark[1],
    Jia Wu\textsuperscript{\rm 2}
}
\affiliations {
    \textsuperscript{\rm 1}School of Computer Science, Wuhan University, Wuhan, China\\
    \textsuperscript{\rm 2}Department of Computing, Macquarie University, Sydney, Australia\\
    \textsuperscript{\rm 3}Wuhan University Shenzhen Research Institute, Shenzhen, China\\
    \{jiameng.chen, yidaxiong, likun98, zhanghongzhi, caixiantao, hwb\}@whu.edu.cn, jia.wu@mq.edu.au
}

\usepackage{bibentry}

\begin{document}

\maketitle

\begin{abstract}
Computational antibody design holds immense promise for therapeutic discovery, yet existing generative models are fundamentally limited by two core challenges: (i)~a lack of dynamical consistency, which yields physically implausible structures, and (ii)~poor generalization due to data scarcity and structural bias. We introduce \textbf{FP-AbDiff}, the first antibody generator to enforce Fokker–Planck Equation (FPE) physics along the entire generative trajectory. Our method minimizes a novel FPE residual loss over the mixed manifold of CDR geometries ($\mathbb{R}^3 \times \mathrm{SO}(3)$), compelling locally-learned denoising scores to assemble into a globally coherent probability flow. This physics-informed regularizer is synergistically integrated with deep biological priors within a state-of-the-art SE(3)-equivariant diffusion framework. Rigorous evaluation on the RAbD benchmark confirms that \textbf{FP-AbDiff establishes a new state-of-the-art}. In de novo CDR-H3 design, it achieves a mean Root Mean Square Deviation of 0.99\,\AA{} when superposing on the variable region, a 25\% improvement over the previous state-of-the-art model, AbX, and the highest reported Contact Amino Acid Recovery of 39.91\%. This superiority is underscored in the more challenging six-CDR co--design task, where our model delivers consistently superior geometric precision, cutting the average full-chain Root Mean Square Deviation by $\sim$15\%, and crucially, achieves the highest full-chain Amino Acid Recovery on the functionally dominant CDR-H3 loop (45.67\%). By aligning generative dynamics with physical laws, FP-AbDiff enhances robustness and generalizability, establishing a principled approach for physically faithful and functionally viable antibody design. Implementation details and code availability are provided in the Supplementary Materials.
\end{abstract}

\section{Introduction}
\label{sec:introduction}
Antibodies are indispensable tools in medicine, where their therapeutic and diagnostic efficacy hinges on the precise structural integrity and conformational dynamics of the complementarity-determining regions (CDRs)~\cite{carter2006potent, north2011new}. Among these, the hypervariable CDR-H3 loop is widely recognized as the principal determinant of binding specificity~\cite{kuroda2012computer}. Despite recent advances, computational antibody design still faces two key unresolved challenges: (i) a lack of dynamical consistency, which leads to unstable structural transitions; and (ii) poor generalization due to limited and biased training datasets, impeding applicability to novel, clinically relevant antigens.
\begin{figure}[t!]
    \centering
    \includegraphics[width=\linewidth]{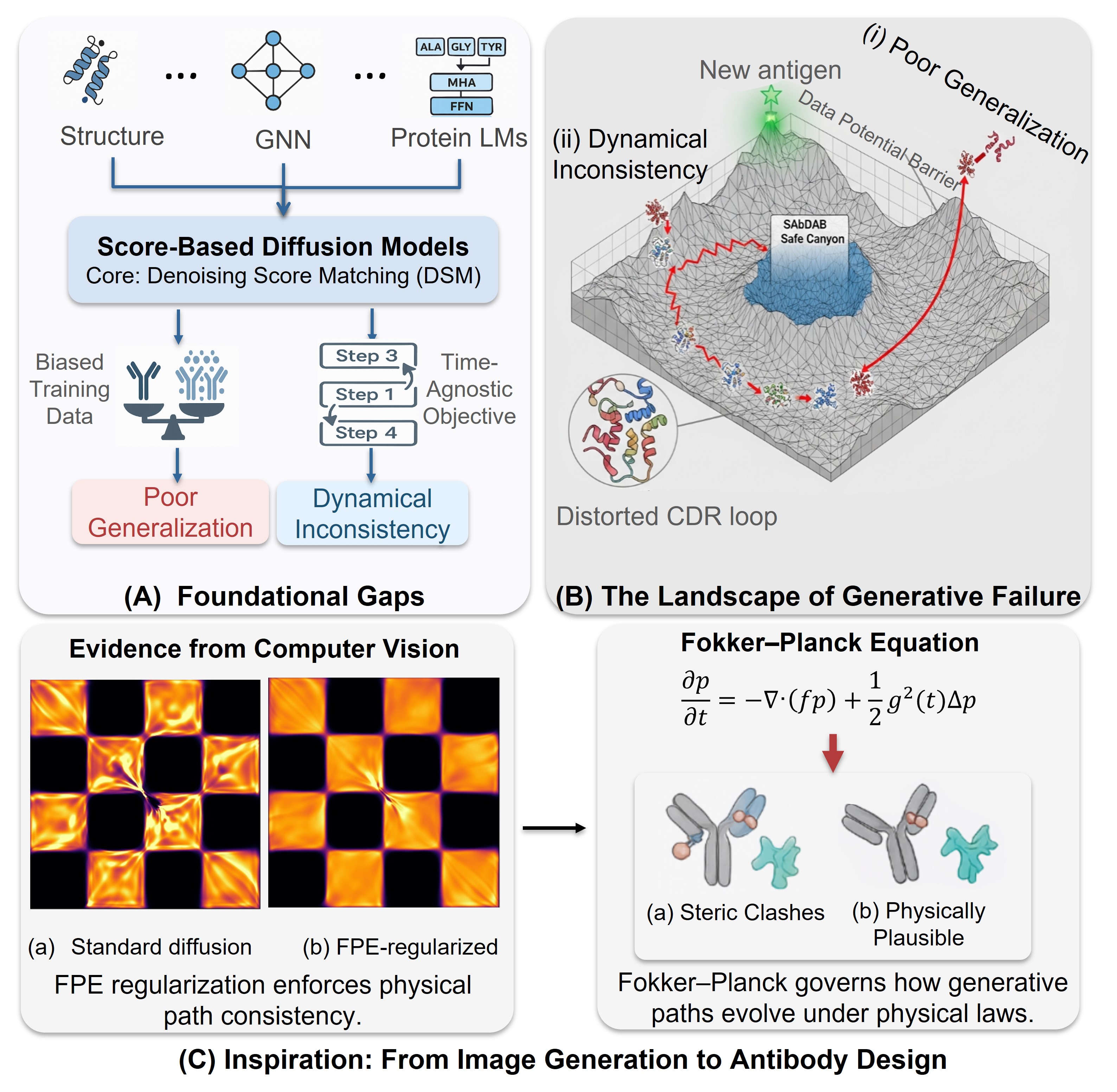}
    \caption{
    Motivating FPE-Regularized Antibody Design: Gaps, Failures, and Inspirations.
    \textbf{(a)}~Core flaws in current models, such as biased data and time-agnostic objectives, lead to \textbf{(b)}~catastrophic failures in generalization and structural integrity. \textbf{(c)}~Inspired by successes in computer vision, we address this by enforcing physical laws through Fokker-Planck Equation regularization to ensure a physically consistent generative path.
    }
    \label{fig:motivation}
\end{figure}

First, dynamical consistency, which is the requirement that a generative trajectory remain physically coherent across time, remains an unresolved challenge in computational antibody design. CDR specificity and affinity arise from subtle, continuous conformational motions rather than isolated structural snapshots \cite{north2011new,hie2024efficient}. Diffusion models such as DiffAb \cite{luo2022antigen} and AbDiffuser \cite{martinkus2023abdiffuser} have reduced backbone RMSD and steric clashes \cite{ruffolo2023fast}, yet they optimize structures at independent noise levels and never constrain the path linking them. Their denoising-score-matching (DSM) objective, which is also used by AbX~\cite{zhu2024antibody} and related frameworks~\cite{song2019generative,ho2020denoising}, captures local gradients but ignores global transitions, often producing chemically implausible loop rearrangements, unstable side-chain packing, and energetically strained conformers that require expensive molecular-dynamics rescue~\cite{ingraham2023illuminating,ausserwoger2022non}. Even architectures with geometric or evolutionary priors, including IgFold \cite{ruffolo2023fast}, GearBind \cite{cai2024pretrainable} and AbMEGD \cite{chen2025antibodydesignoptimizationmultiscale,zhang2024cross}, lack an explicit mechanism to enforce temporal coherence.

Second, diffusion generators falter outside the narrow confines of today’s datasets, limiting their real-world value. This data scarcity challenge is not unique to antibody design; in computer vision, for instance, diffusion models trained on limited data are known to exhibit restricted expressiveness and generate biased outputs~\cite{hur2024expanding}. The problem is critically underscored in our domain, where the main benchmark, SAbDab, contains fewer than 5,000 non-redundant complexes and is heavily biased towards a few human-IgG scaffolds bound to viral epitopes~\cite{dunbar2014sabdab, adolf2018rosettaantibodydesign}. This starkly contrasts with clinical needs for antibodies against diverse topologies like polysaccharides and cryptic viral loops, which lie far beyond the training distribution~\cite{raybould2019five, schneider2022sabdab,elesedy2021provably}. Confronted with out-of-distribution (OOD) tasks~\cite{ijcai2024p234, ijcai2024p235}, state-of-the-art models, including diffusion-based and symmetry-aware graph methods, often revert to familiar motifs and produce paratopes that are sterically or energetically invalid~\cite{ruffolo2023fast, luo2022antigen, martinkus2023abdiffuser, kong2023end}. This brittleness arises from (i) a data bottleneck starving models of CDR diversity and (ii) the absence of global regularization, which over-concentrates probability mass in biased regions. An effective next-generation framework must therefore fuse rich biological priors with a principled constraint that maintains globally consistent probability flow even when data are scarce.

To bridge the dual gaps of dynamical consistency and generalization, we introduce FP-AbDiff, the first antibody generator that enforces Fokker–Planck physics along the entire diffusion path. Building on evidence that trajectories become physically faithful when their score fields satisfy the Fokker–Planck equation (FPE) \cite{lai2023fp,song2019generative}, this principle is extended to the mixed geometry of CDRs. FP-AbDiff couples variance-preserving diffusion in $\mathbb{R}^{3}$ (heavy-atom translations) with variance-exploding diffusion on $SO(3)$ (residue-frame rotations) and embeds the joint FPE residual as a physics-informed regularizer that forces local denoising scores to assemble into a globally coherent probability flow.  
A clean CDR predicted by the network is first converted, via indirect score inference, into exact translational and rotational scores; their residual then drives back-propagation, aligning the trajectory with FPE dynamics. FP-AbDiff inherits key biological priors from AbX \cite{zhu2024antibody}, including ESM-2 evolutionary embeddings, SE(3)-equivariant message passing, and van der Waals and bond geometry terms, while introducing a novel physics-informed residual loss. Together, these components produce CDR trajectories that (i) are dynamically self-consistent, eliminating chemically implausible loop transitions and unstable side-chain packings, and (ii) remain robust on unseen antigen interfaces, thereby reducing costly molecular-dynamics refinement and accelerating therapeutic discovery.

FP-AbDiff is evaluated against state-of-the-art baselines on the RAbD benchmark, confirming that Fokker–Planck consistency provides a clear advantage. Key metrics include Amino Acid Recovery (AAR) and Root Mean Square Deviation (RMSD), assessed over both the full chain ($^{Full}$) and the functional variable region ($^{Fv}$). In de novo CDR-H3 design, FP-AbDiff sets a new benchmark in structural fidelity, achieving a sub-angstrom mean RMSD$^{Fv}$ of 0.99\,\AA{} (a 25\% improvement over AbX) and the highest reported Contact AAR (CAAR) of 39.91\%. This superiority is underscored in the more challenging six-CDR co-design task, where our model achieves superior geometric precision across the entire paratope, delivering the lowest RMSD$^{Full}$ on all six CDRs and the highest AAR$^{Full}$ on the immunologically critical H3 loop. Furthermore, in affinity optimization, FP-AbDiff demonstrates a markedly more stable trajectory. Ablation studies indicate that these multi-faceted performance gains can be directly attributed to the FPE regularizer, which in turn establishes a new benchmark for robust and physically consistent in silico immuno-engineering.

In summary, our contributions are:
\begin{itemize}
    \item FP-AbDiff introduces the first CDR-targeted diffusion framework that enforces score–Fokker–Planck consistency over $\mathbb{R}^3 \times \mathrm{SO}(3)$, ensuring globally coherent probability flows and eliminating non-physical loop transitions.
    
    \item It unifies Fokker–Planck physics with evolutionary, geometric, and energetic priors into a single objective, enabling dynamically consistent and generalizable antibody generation.
    
    \item Extensive evaluations confirm that FP-AbDiff achieves state-of-the-art performance across antibody design and optimization tasks, demonstrating the broad benefit of physics-informed regularization.
\end{itemize}

\section{Related Work}

Computational antibody design has rapidly evolved from classical, energy-based methods to structure-aware generative frameworks. Early approaches, exemplified by methods like RosettaAntibodyDesign and AbDesign~\cite{adolf2018rosettaantibodydesign, kuroda2012computer, ruffolo2021deciphering}, relied on statistical energy functions and Monte Carlo sampling but were hampered by prohibitive computational costs and limited sampling efficacy over the vast conformational space. The emergence of protein language models~\cite{elnaggar2007prottrans, shin2021protein} enabled efficient, sequence-centric generation, treating proteins as textual inputs~\cite{xiong2025textguidedmultipropertymolecularoptimization}. However, such models often neglect spatial and geometric priors critical for antibody-antigen binding.

Recent advances have introduced geometric and equivariant models for joint sequence–structure design. Autoregressive methods (HERN~\cite{jin2022antibody}) and GNN-based models (MEAN, dyMEAN~\cite{kong2022conditional,kong2023end}) have shown strong performance in CDR co-design, while predictors like AlphaFold2~\cite{jumper2021highly} and SE(3)-Fold~\cite{norman2020thermometry,ahdritz2024openfold} yield high-fidelity structures but lack generative capacity. Diffusion-based models such as DiffAb~\cite{luo2022antigen} and AbDiffuser~\cite{martinkus2024abdiffuser} mark progress, yet often diverge from principled score-based formulations. AbX~\cite{zhu2024antibody} remains the only model to learn a continuous-time score field. However, all existing approaches, discrete or continuous, remain time-agnostic and lack dynamical consistency. Drawing on Fokker–Planck regularization~\cite{lai2023fp}, we present the first framework to impose physical self-consistency across the entire diffusion trajectory in antibody generation.

\begin{figure*}[!t]
    \centering
    \includegraphics[width=0.95\textwidth]{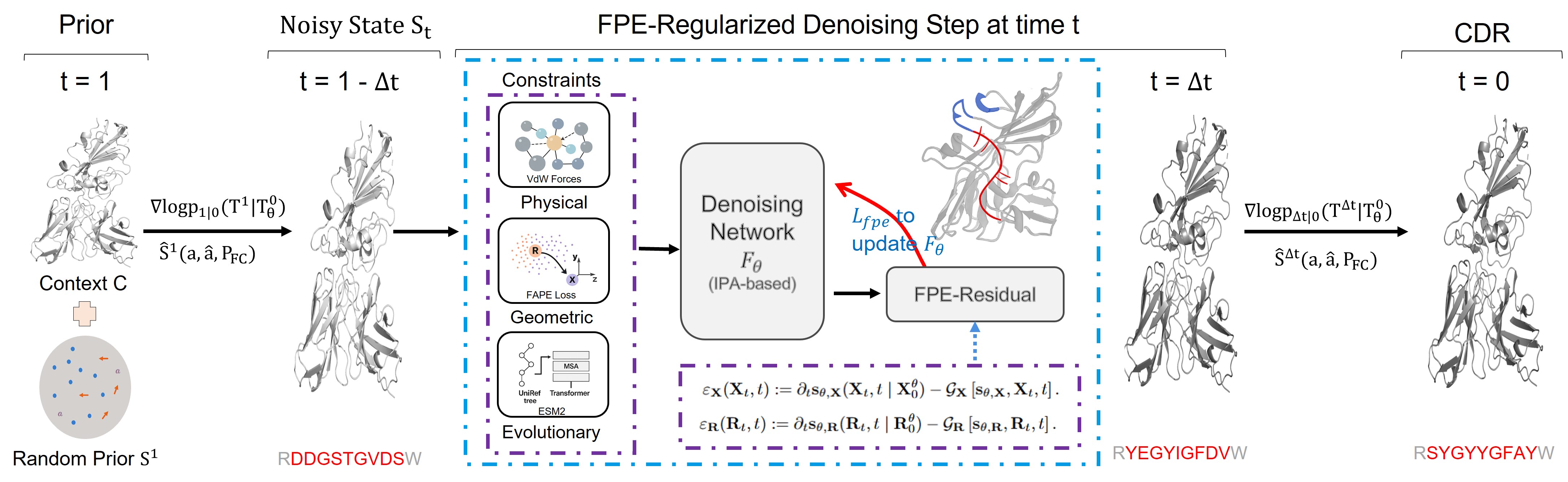}
    \caption{
    Overview of FP-AbDiff. FP-AbDiff leverages a Continuous Time Markov Chain (CTMC) for CDR sequence modeling and a score-based diffusion framework for CDR structure generation. It incorporates physical and geometric constraints via a physics-informed loss derived from the Fokker–Planck Equation and applies evolutionary priors within the model architecture. The grey regions indicate the antigen and antibody framework, while the red regions highlight the designed CDRs in the antibody.
    }
    
    \label{fig:overview}
\end{figure*}

\section{Methods}
\label{sec:methods}

FP-AbDiff is a unified generative framework for antibody design that integrates stochastic dynamics with physics-informed regularization. As shown in Figure~\ref{fig:overview}, it jointly models discrete sequence and continuous structure spaces using domain-specific dynamics: a context-conditioned Continuous Time Markov Chain (CTMC)~\cite{zhu2024antibody, campbell2022continuous} for sequences, and stochastic differential equations (SDEs) for structures. Physical consistency is enforced through a Fokker–Planck residual regularizer, introduced in Section~\ref{sec:fpe_residual_bridge} and efficiently implemented in Section~\ref{sec:implementation_details}. A unified training objective is then defined (Section~\ref{sec:training}), and reverse-time sampling is deployed to generate candidate antibodies (Section~\ref{sec:sampling_algorithm}).

\subsection{Problem Formulation and Notations}
\label{sec:problem_formulation}
Antibody design is formulated as the conditional generation of a CDR given its structural context~$\mathcal{C}$ (antigen and framework). A CDR of $N_{\text{CDR}}$ residues is defined by its ground-truth state at $t=0$, $\mathcal{S}_0 = (\mathbf{A}_0, \mathbf{X}_0, \mathbf{R}_0)$, which comprises: (1)~the amino acid sequence $\mathbf{A}_0 = (a_1, \dots, a_{N_{\text{CDR}}})$ with $a_i \in \{\text{ACDEFGHIKLMNPQRSTVWY}\}$; (2)~the heavy atom coordinates $\mathbf{X}_0 \in \mathbb{R}^{D_X}$, where $D_X = 3 \times \sum n_i$; and (3)~the residue orientations $\mathbf{R}_0 \in \mathrm{SO}(3)^{N_{\text{CDR}}}$. For brevity, the geometric state is denoted as ${T}_0 := (\mathbf{X}_0, \mathbf{R}_0)$. The goal is to learn the conditional distribution $p(\mathcal{S}_0 \mid \mathcal{C})$ via a score-based diffusion framework. The forward process perturbs the full state $\mathcal{S}_0$ into a noisy state $\mathcal{S}_t = (\mathbf{A}_t, \mathbf{X}_t, \mathbf{R}_t)$ over time $t \in [0,1]$, where ${T}_t := (\mathbf{X}_t, \mathbf{R}_t)$ is the corresponding noisy geometric state. The score network then learns to reverse this process by predicting the denoised geometric state ${T}_0^\theta$ from ${T}_t$, thereby implicitly defining a time-dependent score field $\mathbf{s}_\theta({T}_t, t)$. All processes are implicitly conditioned on $\mathcal{C}$.

\subsection{Stochastic Dynamics of CDR Structures}
\label{sec:sde_framework}

CDR structure generation is modeled as a reverse-time diffusion over $(\mathbf{X}_t, \mathbf{R}_t)$ on the manifold $\mathbb{R}^{D_X} \times \mathrm{SO}(3)^{N_{\text{CDR}}}$, governed by SDEs~\cite{majumdar2023introduction,hsu2002stochastic}. To enforce dynamic consistency, a Fokker–Planck–derived regularizer is introduced to constrain the evolution of the score field.

\subsubsection{Forward Diffusion Processes on the Structure Manifold}
\label{sec:sde_forward_process}

To handle the mixed geometry of CDRs, translational and rotational components are governed by separate SDEs.

\paragraph{Translational Dynamics in Euclidean Space.}
The backbone coordinates $\mathbf{X}_t \in \mathbb{R}^{3}$ evolve according to a variance-preserving (VP) SDE, specifically an Ornstein–Uhlenbeck process \cite{song2021scorebased}:

\begin{equation}
    d\mathbf{X}_t = -\frac{1}{2}\beta_X(t)\mathbf{X}_t \, dt + \sqrt{\beta_X(t)} \, d\mathbf{W}_{\mathbf{X},t}.
    \label{eq:sde_r3_forward}
\end{equation}
where $\beta_X(t) > 0$ is a predefined noise schedule and $\mathbf{W}_{\mathbf{X},t}$ denotes standard Brownian motion. This SDE induces a time-dependent Gaussian transition kernel:
\begin{equation}
    p_{t|0}(\mathbf{X}_t|\mathbf{X}_0) = \mathcal{N}(\mathbf{X}_t ; \alpha_X(t)\mathbf{X}_0, \sigma_X^2(t)\mathbf{I}).
    \label{eq:sde_r3_kernel}
\end{equation}
with coefficients determined by the integrated noise: 
$\alpha_X(t) = e^{-\frac{1}{2}\bar{\beta}_X(t)}$, $\sigma_X^2(t) = 1 - \alpha_X^2(t)$, where $\bar{\beta}_X(t) = \int_0^t \beta_X(s) \, ds$.  As \( t \to 1 \), the distribution of \( \mathbf{X}_t \) converges to the standard normal prior \( \mathcal{N}(\mathbf{0}, \mathbf{I}) \).

\paragraph{Rotational Dynamics on the $SO(3)$ Manifold.}
Each residue’s orientation $\mathbf{R}_{i,t} \in SO(3)$ evolves via a variance-exploding (VE) SDE describing Brownian motion on the Lie group \cite{nikolayev1997normal,leach2022denoising}:
\begin{equation}
    d\mathbf{R}_{i,t} = \sqrt{\beta_R(t)} \sum_{a=1}^{3} (\mathbf{R}_{i,t} E_a) \circ dW^a_t.
    \label{eq:sde_so3_forward}
\end{equation}
where $\{E_a\}_{a=1}^3$ is an orthonormal basis of the Lie algebra $\mathfrak{so}(3)$, $\beta_R(t) > 0$ is the time-dependent diffusion rate, and $\circ$ denotes Stratonovich integration. The transition kernel is the isotropic Gaussian on $SO(3)$:
$p_{\mathrm{IGSO}(3)}(\mathbf{R}_{i,0}^\top \mathbf{R}_{i,t}; \sigma_R^2(t)), \quad \sigma_R^2(t) = \int_0^t \beta_R(s) \, ds.$ As $t \to 1$, it converges to the Haar (uniform) distribution on $SO(3)$.

\subsubsection{Governing Equations: The FPE and Score Dynamics}
\label{sec:fpe_governing_equations}

While the forward SDEs define stochastic sample trajectories, the evolution of their densities is governed by the FPE. For a general SDE, $d\mathbf{x}_t = \mathbf{f}(\mathbf{x}_t, t) dt + g(t) d\mathbf{W}_t,$ the FPE describes how the probability density $p(\mathbf{x}, t)$ evolves:
\begin{equation}
    \frac{\partial p}{\partial t} = -\nabla \cdot (\mathbf{f} p) + \frac{1}{2} g^2(t) \Delta p.
    \label{eq:fpe_general}
\end{equation}
Here, $\mathbf{f}$ is the drift field, $g(t)$ controls diffusion intensity, and $\Delta p$ denotes the Laplacian capturing isotropic spreading.

\paragraph{Dynamics on Euclidean Space ($\mathbb{R}^{D_X}$).}
The FPE corresponding to the translational SDE (Eq.~\ref{eq:sde_r3_forward}) is a partial differential equation for the density $p(\mathbf{X},t)$:
\begin{equation}
    \frac{\partial p}{\partial t} = \frac{1}{2} \beta_X(t) \nabla_{\mathbf{X}} \cdot (\mathbf{X} p) + \frac{1}{2} \beta_X(t) \Delta_{\mathbf{X}} p.
    \label{eq:fpe_r3}
\end{equation}
where $\nabla_{\mathbf{X}}$ denotes the spatial gradient, $\operatorname{div}_{\mathbf{X}}$ the divergence, and $\Delta_{\mathbf{X}}$ the Laplacian operator.
From this, the evolution for the log-density $\ell_{\mathbf{X}} = \log p(\mathbf{X},t)$ is derived. Letting $\mathbf{s}_{\mathbf{X}} = \nabla_{\mathbf{X}} \ell_{\mathbf{X}}$, the temporal evolution $\frac{\partial \ell_{\mathbf{X}}}{\partial t}$ is given by the operator $\mathcal{L}_X[\mathbf{s}_{\mathbf{X}}]$:
\begin{equation}
    \mathcal{L}_X[\mathbf{s}_{\mathbf{X}}] := \frac{1}{2}\beta_X(t)\left[ D_X + \mathbf{X}\cdot\mathbf{s}_{\mathbf{X}} + \operatorname{div}_{\mathbf{X}}(\mathbf{s}_{\mathbf{X}}) + \|\mathbf{s}_{\mathbf{X}}\|^2\right].
    \label{eq:log_density_evolution_r3}
\end{equation}

The Score FPE is obtained by taking the gradient of this entire expression, $\partial_t \mathbf{s}_{\mathbf{X}} = \nabla_{\mathbf{X}} \mathcal{L}_X[\mathbf{s}_{\mathbf{X}}]$. This yields the final evolution operator $\mathcal{G}_{\mathbf{X}}$:
\begin{equation}
    \mathcal{G}_{\mathbf{X}}[\mathbf{s}_{\mathbf{X}}, \mathbf{X}, t] := \frac{1}{2}\beta_X(t)\left[ \mathbf{s}_{\mathbf{X}} + (\nabla_{\mathbf{X}}\mathbf{s}_{\mathbf{X}})\mathbf{X} + \mathbf{H}_{\mathbf{X}}(\mathbf{s}_{\mathbf{X}}) \right],
    \label{eq:score_fpe_r3_op}
\end{equation}
where the higher-order derivative term is compactly defined as:
\begin{equation}
    \mathbf{H}_{\mathbf{X}}(\mathbf{s}_{\mathbf{X}}) := \nabla_{\mathbf{X}}\left(\operatorname{div}_{\mathbf{X}}(\mathbf{s}_{\mathbf{X}}) + \|\mathbf{s}_{\mathbf{X}}\|^2\right).
    \label{eq:h_term_r3}
\end{equation}
This equation provides the precise, deterministic rule for how the coordinate score field $\mathbf{s}_{\mathbf{X}}$ must evolve over time.

\paragraph{Dynamics on the $SO(3)$ Manifold.}
A parallel derivation on the $SO(3)$ manifold starts with its specific FPE, which involves the Laplace-Beltrami operator $\Delta_{SO(3)}$:
\begin{equation}
    \frac{\partial p}{\partial t} = \frac{1}{2}\beta_R(t)\Delta_{SO(3)}p(R,t).
    \label{eq:fpe_so3}
\end{equation}
The corresponding log-density evolution for the Riemannian score $\mathbf{s}_{\mathbf{R}} = \nabla_{SO(3)} \log p$ is:
\begin{equation}
    \frac{\partial \ell_{\mathbf{R}}}{\partial t} = \mathcal{L}_{R}[\mathbf{s}_{\mathbf{R}}] := \frac{1}{2}\beta_R(t) \left( \operatorname{div}_{SO(3)} \mathbf{s}_{\mathbf{R}} + \|\mathbf{s}_{\mathbf{R}}\|_g^2 \right),
    \label{eq:log_density_evolution_so3}
\end{equation}

The Score FPE is obtained by taking the Riemannian gradient of the log-density evolution: $\partial_t \mathbf{s}_{\mathbf{R}} = \nabla_{SO(3)} \mathcal{L}_R[\mathbf{s}_{\mathbf{R}}]$. Leveraging the Weitzenböck identity and the known Ricci curvature of $SO(3)$ under its canonical metric \cite{debortoli2022riemannianscorebasedgenerativemodelling,lai2023fp}, the closed-form evolution operator $\mathcal{G}_{\mathbf{R}}$ is derived. This operator can be expressed compactly by defining a higher-order term $\mathbf{H}_{\mathbf{R}}(\mathbf{s}_{\mathbf{R}}) := 2 (\nabla_{SO(3)} \mathbf{s}_{\mathbf{R}})^\top \mathbf{s}_{\mathbf{R}}$, yielding:
\begin{equation}
    \mathcal{G}_{\mathbf{R}}[\mathbf{s}_{\mathbf{R}}, \mathbf{R}, t] := \frac{1}{2} \beta_R(t) \left[ \Delta_B \mathbf{s}_{\mathbf{R}} - 2\mathbf{s}_{\mathbf{R}} + \mathbf{H}_{\mathbf{R}}(\mathbf{s}_{\mathbf{R}}) \right].
    \label{eq:score_fpe_so3_op}
\end{equation}

\subsection{Bridging Theory and Prediction via the FPE Residual} 
\label{sec:fpe_residual_bridge}

To enforce dynamical consistency, the score field $\mathbf{s}_\theta$, implicitly inferred from the network's denoising prediction ${T}_0^\theta = (\mathbf{X}_0^\theta, \mathbf{R}_0^\theta)$, is regularized. Leveraging the theoretical link between score matching and denoising~\cite{vincent2011connection, song2021scorebased}, and the known transition kernels of the forward SDEs (Eqs.~\ref{eq:sde_r3_kernel} and \ref{eq:sde_so3_forward}), the model-implied scores are analytically derived as:

For the translational component, this analytical relationship yields:
\begin{equation}
    \mathbf{s}_{\theta,\mathbf{X}}(\mathbf{X}_t,t \mid \mathbf{X}_0^\theta) = -\frac{\mathbf{X}_t - \alpha_X(t)\mathbf{X}_0^\theta}{\sigma_X^2(t)}.
    \label{eq:indirect_score_r3}
\end{equation}
For the rotational component, the same principle on the $SO(3)$ manifold gives the Riemannian score:
\begin{equation}
    \mathbf{s}_{\theta,\mathbf{R}}(\mathbf{R}_t, t \mid \mathbf{R}_0^\theta) = \nabla_{SO(3)}\log p_{\mathrm{IGSO}(3)}\big((\mathbf{R}_0^\theta)^\top \mathbf{R}_t; \sigma_R^2(t)\big).
    \label{eq:indirect_score_so3}
\end{equation}

The Fokker–Planck residual $\boldsymbol{\varepsilon}$ is defined as the deviation between the temporal derivative of this score field and the evolution dictated by the Score-FPE operator $\mathcal{G}$. This parallels residuals in fluid dynamics \cite{milnor1968note}, such as those from the Navier–Stokes equations, which quantify deviation from conservation laws.

For translational dynamics in $\mathbb{R}^{D_X}$, the residual is a vector field:
\begin{equation}
    \boldsymbol{\varepsilon}_{\mathbf{X}}(\mathbf{X}_t,t) := \partial_t \mathbf{s}_{\theta,\mathbf{X}}(\mathbf{X}_t,t \mid \mathbf{X}_0^\theta)
    - \mathcal{G}_{\mathbf{X}}\left[\mathbf{s}_{\theta,\mathbf{X}}, \mathbf{X}_t, t\right].
    \label{eq:residual_r3}
\end{equation}

For rotational dynamics on $SO(3)$, it becomes a tangent vector field:
\begin{equation}
    \boldsymbol{\varepsilon}_{\mathbf{R}}(\mathbf{R}_t,t) := \partial_t \mathbf{s}_{\theta,\mathbf{R}}(\mathbf{R}_t,t \mid \mathbf{R}_0^\theta)
    - \mathcal{G}_\mathbf{R}\left[\mathbf{s}_{\theta,\mathbf{R}}, \mathbf{R}_t, t \right].
    \label{eq:residual_so3}
\end{equation}

Minimizing $\|\boldsymbol{\varepsilon}\|^2$ enforces physical alignment between the generative flow and the underlying SDE-based dynamics, without interfering with the prediction objective.

\subsection{Efficient Implementation and Complexity Analysis}
\label{sec:implementation_details}

The FPE residual is implemented with stable and differentiable approximations. All computations are conditioned on the denoising prediction ${T}_0^\theta$, from which the score field $\mathbf{s}_\theta$ over noisy inputs ${T}_t$ is analytically derived.

\paragraph{Temporal Derivative.}
The time derivative $\partial_t \mathbf{s}_\theta({T}_t, t)$ is estimated via a second-order central finite-difference scheme. For a given time step $\delta t$, the network is queried at $t_\pm = t \pm \delta t$ to obtain denoised states ${T}_{0,\pm}^\theta$ and their corresponding scores $\mathbf{s}_{\theta,\pm}$, yielding the approximation:
\begin{equation}
    \partial_t \mathbf{s}_\theta({T}_t, t) 
    \approx \frac{ \mathbf{s}_{\theta,+} - \mathbf{s}_{\theta,-} }{2\delta t}.
    \label{eq:time_deriv_finite_diff}
\end{equation}

\paragraph{Spatial Operators.}
The spatial operators $\mathcal{G}_{\mathbf{X}}$ and $\mathcal{G}_{\mathbf{R}}$ are evaluated using two key approximations for computational efficiency. First, divergence terms are stably estimated via Hutchinson's trick, using a single Rademacher probe vector, and normalized by the number of atoms. Second, a first-order Euclidean approximation is used for all manifold derivatives within the tangent space at the identity, a standard and accurate strategy for small training steps~\cite{yim2023se, debortoli2022riemannianscorebasedgenerativemodelling}.

\paragraph{Computational Complexity.}
The score network scales as \(\mathcal{O}(L N^2)\) with CDR length \(N\) and \(L\) IPA layers. The FPE regularizer's overhead is analyzed in terms of forward (F) and backward (B) passes. It adds two extra F passes for the temporal derivative and two lightweight spatial steps that bypass the main IPA layers. Consequently, the total cost per step increases from \(\mathcal{O}(F + B)\) to \(\mathcal{O}(3F + B)\) plus negligible spatial overhead, which, due to the two additional forward passes required for the temporal derivative, empirically results in a modest ~8\% increase in wall-clock training time.

\subsection{Training Objectives}
\label{sec:training}


\subsubsection{Fidelity Losses}
\label{sec:fidelity_losses}
The fidelity loss encourages the model to reconstruct realistic structure and sequence.
\paragraph{Translational Score Matching.} Equivalent to VP-SDE denoising score matching~\cite{vincent2011connection}, the translational loss on Cartesian coordinates is:
\begin{equation}
    \mathcal{L}_{\text{DSM}}^X = \frac{1}{N_{\text{CDR}}} \sum_{i=1}^{N_{\text{CDR}}} \left\| \mathbf{x}_i^0 - (\mathbf{x}_i^0)^\theta \right\|^2,
\end{equation}
where $(\mathbf{x}_i^0)^\theta$ is the predicted clean coordinate for residue $i$.

\paragraph{Rotational Score Matching.} 
For rotation, a Frobenius-norm loss in \(\mathrm{SO}(3)\) is adopted following~\cite{yim2023se}:
\begin{equation}
    \mathcal{L}_{\text{DSM}}^R = \frac{1}{N_{\text{CDR}}} \sum_{i=1}^{N_{\text{CDR}}} \left\| \text{Exp}(\mathbf{s}_{\text{true}}^i) - \text{Exp}(\mathbf{s}_{\theta,i}^R) \right\|_F^2,
\end{equation}
where $\mathbf{s}_{\text{true}}^i$ and $\mathbf{s}_{\theta,i}^R$ are the ground-truth and predicted scores in the Lie algebra $\mathfrak{so}(3)$~\cite{sola2018micro}; $\text{Exp}(\cdot)$ maps to $SO(3)$, and $\|\cdot\|_F$ is the Frobenius norm.

\paragraph{Total Fidelity Loss.} The overall fidelity objective combines structure and sequence components:
\begin{equation}
    \mathcal{L}_{\text{fid}} = \mathcal{L}_{\text{DSM}}^X + \mathcal{L}_{\text{DSM}}^R + 0.4 \cdot \mathcal{L}_{\text{CE}},
\end{equation}
where \(\mathcal{L}_{\text{CE}}\) is the standard masked cross-entropy over amino acid types.

\subsubsection{Biophysical Plausibility Priors}
\label{sec:priors}

To encourage biochemically plausible conformations, a set of biophysical priors, $\mathcal{L}_{\text{priors}}$, adapted from AlphaFold2~\cite{jumper2021highly}, is introduced. These priors act on the predicted final structure $\mathcal{S}_0^\theta$ at early denoising steps ($t < \tau$), refining fine-grained geometry through both geometric and physical constraints.

The core geometric prior is the Frame Aligned Point Error ($\mathcal{L}_{\text{FAPE}}$), which penalizes SE(3)-invariant deviations between predicted and ground-truth structures. Additional priors include $\mathcal{L}_{\text{dist}}$ (a distogram loss), $\mathcal{L}_{\text{pLDDT}}$ (a confidence prediction loss), $\mathcal{L}_{\text{viol}}$ (a penalty for bond violations and steric clashes), and a backbone atom refinement loss ($\mathcal{L}_{bb}$) that directly supervises backbone heavy-atom positions at low noise levels ($t < 0.25$).

The total prior loss is a weighted sum:
\begin{equation}
\begin{split}
    \mathcal{L}_{\text{priors}} ={}& \mathcal{L}_{\text{FAPE}} + 0.5 \mathcal{L}_{\text{dist}} + 0.1 \mathcal{L}_{\text{pLDDT}} \\
    & + 0.03 \mathcal{L}_{\text{viol}} + 0.25 \mathcal{L}_{bb}.
\end{split}
\label{eq:loss_priors_total}
\end{equation}

\subsubsection{Dynamical Consistency Regularizer}
\label{sec:fpe_regularizer}
To ensure physical consistency during generation, an FPE-based loss $\mathcal{L}_{\text{fpe}}$ is introduced, defined as the expected, dimension-normalized, and time-weighted squared norm of the residual $\boldsymbol{\varepsilon}$:
\begin{equation}
    \mathcal{L}_{\text{fpe}}(\theta) = \mathbb{E}_{t, \mathcal{S}_t} \left[
    w(t) \left( \frac{\|\boldsymbol{\varepsilon}_{\mathbf{X}}\|^2}{D_X} + \frac{\|\boldsymbol{\varepsilon}_{\mathbf{R}}\|^2}{D_R} \right)
    \right],
    \label{eq:loss_fpe}
\end{equation}
where $\boldsymbol{\varepsilon}_{\mathbf{X}}$ and $\boldsymbol{\varepsilon}_{\mathbf{R}}$ are the translational and rotational FPE residuals (see Eqs.~\ref{eq:residual_r3},~\ref{eq:residual_so3}), and $D_X$, $D_R$ denote their respective dimensions. The weight $w(t)$ balances contributions across time steps.

\subsubsection{Complete Loss Function}
\label{sec:complete_loss}
The final loss combines all components, with biophysical priors applied only to near-denoised structures ($t < \tau$):
\begin{equation}
    \mathcal{L}_{\text{total}} = \mathcal{L}_{\text{fid}} + I_{t < \tau} \mathcal{L}_{\text{priors}} + 0.05 \cdot \mathcal{L}_{\text{fpe}}.
    \label{eq:loss_total}
\end{equation}

\subsection{Sampling Algorithm}
\label{sec:sampling_algorithm}
To generate novel CDRs, reverse-time stochastic dynamics of both structure and sequence are simulated in discrete steps of size \(\tau\). Structural coordinates, including translation and rotation, are updated using Euler–Maruyama discretization of the reverse-time SDE~\cite{yim2023se}. In parallel, sequences are sampled via a tau-leaping scheme that approximates reverse-time CTMC dynamics by aggregating Poisson-distributed mutation events over \(\tau\)~\cite{gillespie2001approximate}:
\begin{equation}
    \mathbf{A}^{t - \tau} = \mathbf{A}^t + \sum_{d=1}^{N_{\text{CDR}}} \sum_{s \ne a_d^t} P_s (s - a_d^t)\mathbf{e}_d,
\end{equation}
where \(\mathbf{e}_d\) is a one-hot vector at position \(d\), and \(P_s \sim \text{Poisson}(\tau \cdot \hat{S}_t^\theta(a^t, s))\) denotes the number of mutations to amino acid \(s\) under the predicted transition rates \(\hat{S}_t^\theta\), computed by a dedicated sequence head. Final side-chains are built via a rotamer library~\cite{misura2004analysis} and refined using Rosetta FastRelax~\cite{adolf2018rosettaantibodydesign} to resolve steric clashes.

\section{Experiments}
\label{sec:experiments}
FP-AbDiff is evaluated under the central hypothesis that enforcing FPE consistency improves generative accuracy and physical plausibility. Section~\ref{sec:co_design} assesses de novo sequence–structure co-design. Section~\ref{sec:affinity} addresses affinity optimization. Section~\ref{sec:ablation} isolates the impact of the FPE regularizer via ablation.

\subsection{Sequence and Structure Co-design}
\label{sec:co_design}

The primary evaluation of FP-AbDiff targets de novo co-generation of CDR sequences and backbone structures on the RAbD benchmark of 60 antibody–antigen complexes~\cite{adolf2018rosettaantibodydesign}. To assess performance across generative complexity, two tasks are considered: (i) targeted CDR-H3 generation—a canonical benchmark for epitope-specific binding due to its high variability and immunological significance; and (ii) full-paratope design via simultaneous generation of all six CDRs, requiring global structural coherence across spatially distant regions. The latter poses a substantially harder challenge, where our FPE-regularized model provides a key advantage. Training uses a non-redundant SAbDab-derived set~\cite{dunbar2014sabdab} (September 2024), with $\leq$ 40\% CDR-H3 sequence identity between splits to prevent leakage.

\paragraph{Baseline models.}
FP-AbDiff is compared against state-of-the-art baselines spanning key paradigms in antibody design: (i) diffusion models (DiffAb~\cite{luo2022antigen}, AbX~\cite{zhu2024antibody}); (ii) energy-guided pipelines (RosettaAb~\cite{adolf2018rosettaantibodydesign}); (iii) equivariant GNNs modeling residue–structure geometry (dyMEAN~\cite{kong2023end}, MEAN~\cite{kong2022conditional}); and (iv) autoregressive sequence models (HERN~\cite{jin2022antibody}). These baselines provide a comprehensive comparison across neural/non-neural, geometry-aware, and energy-driven approaches.

\paragraph{Evaluation metrics.}
We adopt a unified suite of metrics covering sequence recovery, structural accuracy, and functional viability. To resolve alignment ambiguities, both full-chain ($^{\text{Full}}$) and Fv-region ($^{\text{Fv}}$) evaluations are reported. Sequence recovery is measured via AAR$^{\text{Fv}}$ and AAR$^{\text{Full}}$ (\%), and interface-specific CAAR (\%)~\cite{ramaraj2012antigen}. Structure accuracy includes RMSD$^{\text{Fv}}$ and RMSD$^{\text{Full}}$ (\AA), TM-score~\cite{zhang2004scoring,xu2010significant}, and lDDT~\cite{mariani2013lddt}. Functional viability is assessed via IMP (\% of samples with $\Delta\Delta G < 0$ from Rosetta InterfaceAnalyzer) and DockQ~\cite{basu2016dockq}. All metrics are averaged over 100 generated samples per test complex.

\begin{table}[t]
\centering
\small 
\setlength{\tabcolsep}{1pt} 

\begin{tabular}{lcccccc}
\hline
& \multicolumn{3}{c}{\textbf{Generation}} & \multicolumn{3}{c}{\textbf{Docking}} \\
\cline{2-4} \cline{5-7} 
\textbf{Model} & \textbf{AAR↑} & \textbf{TMscore↑} & \textbf{lDDT↑} & \textbf{CAAR↑} & \textbf{RMSD↓} & \textbf{DockQ↑} \\
\hline
RosettaAb     & 32.31\%          & 0.9717          & 0.8272          & 14.58\%          & 17.70          & 0.137 \\
DiffAb        & 35.31\%          & 0.9695          & 0.8281          & 22.17\%          & 23.24          & 0.158 \\
MEAN          & 37.38\%          & 0.9688          & 0.8252          & 24.11\%          & 17.30          & 0.162 \\
HERN            & 32.65\%          & -               & -               & 19.27\%          & 9.15           & 0.294 \\
dyMEAN          & 43.65\%          & 0.9726          & 0.8454          & 28.11\%          & 8.11           & 0.409 \\
AbX             & \textbf{84.90\%} &\underline{0.9906}         & \textbf{0.9407} & \underline{39.08\%}         & \underline{1.32}          & \underline{0.429} \\
FP-AbDiff & \underline{83.65\%}         & \textbf{0.9929} & \underline{0.9363}        & \textbf{39.91\%} & \textbf{0.99}  & \textbf{0.444} \\
\hline
\end{tabular}

\caption{
    Epitope-binding CDR-H3 design on RAbD.
    Key metrics AAR (\%) and RMSD (\AA) are reported for the Fv-region ($^{Fv}$).
    Best results are in \textbf{bold}; second-best are \underline{underlined}
}
\label{tab:h3_design_comparison}
\end{table}

\paragraph{Experimental Results.}
In the foundational task of epitope-binding CDR-H3 design, FP-AbDiff establishes a new state-of-the-art across sequence, structure, and docking (Table~\ref{tab:h3_design_comparison}). While its overall sequence recovery (AAR$^{\text{Fv}}$) is highly competitive with AbX, FP-AbDiff achieves the top CAAR, indicating a sharper focus on functionally critical, paratope-facing residues. This precision is mirrored in its structural fidelity, where FP-AbDiff not only yields the lowest RMSD$^{\text{Fv}}$ (a 25\% error reduction over AbX) but also attains the highest TM-score. This comprehensive superiority culminates in leading functional viability, evidenced by the best DockQ score. The convergence of top performance across these disparate metrics suggests that enforcing Fokker-Planck consistency unifies the generative process across sequence, geometry, and energetics. This advantage is not localized to CDR-H3. A detailed, per-CDR dissection (Table~\ref{tab:per_cdr_comparison}) reveals that this superiority is uniform across the entire paratope. FP-AbDiff delivers the lowest RMSD$^{\text{Full}}$ for all six CDRs, cutting the geometric error versus AbX by an average of $\sim$15\% (ranging from 9.6\% on H3 to an exceptional 24.4\% on L2). Sequence recovery is similarly strong, topping the two immunologically most critical loops (H3).

\begin{table}[t]
\centering
\small
\setlength{\tabcolsep}{2.1pt} 

\begin{tabular}{llcccccc}
\hline
\textbf{Metric} & \textbf{Method} & \textbf{H1} & \textbf{H2} & \textbf{H3} & \textbf{L1} & \textbf{L2} & \textbf{L3} \\
\hline
\multirow{4}{*}{\textbf{AAR(\%)↑}} & DiffAb    & 70.01          & 38.52          & 28.05          & 61.07          & 58.58          & 47.57 \\
                                & dyMEAN    & 75.71          & \underline{68.48}          & 37.50          & 75.55          & \underline{83.09}    & 52.11 \\
                                & AbX       & \underline{80.72}    & \textbf{70.73} & \underline{45.18}    & \textbf{79.37} & \textbf{84.53} & \textbf{65.92} \\
                                & FP-AbDiff & \textbf{81.67} & 66.91    & \textbf{45.67} & \underline{77.09}    & 77.64          & \underline{58.48} \\
\hline
\multirow{4}{*}{\textbf{RMSD(\AA)↓}} & DiffAb    & 0.88           & 0.78     & 2.86           & 0.85           & 0.55           & 1.39 \\
                                & dyMEAN    & 1.09           & 1.11           & 3.88           & 1.03           & 0.66           & 1.44 \\
                                & AbX       & \underline{0.85}     & \underline{0.76}           & \underline{2.50}     & \underline{0.78}     & \underline{0.45}     & \underline{1.18} \\
                                & FP-AbDiff & \textbf{0.74}  & \textbf{0.64}  & \textbf{2.26}  & \textbf{0.68}  & \textbf{0.34}  & \textbf{0.99} \\
\hline
\end{tabular}

\caption{
    Per-CDR performance on the RAbD test set. All metrics are calculated on the full chain ($^{Full}$).
}
\label{tab:per_cdr_comparison}
\end{table}

\begin{figure}[h]
    \centering
    \includegraphics[width=\linewidth]{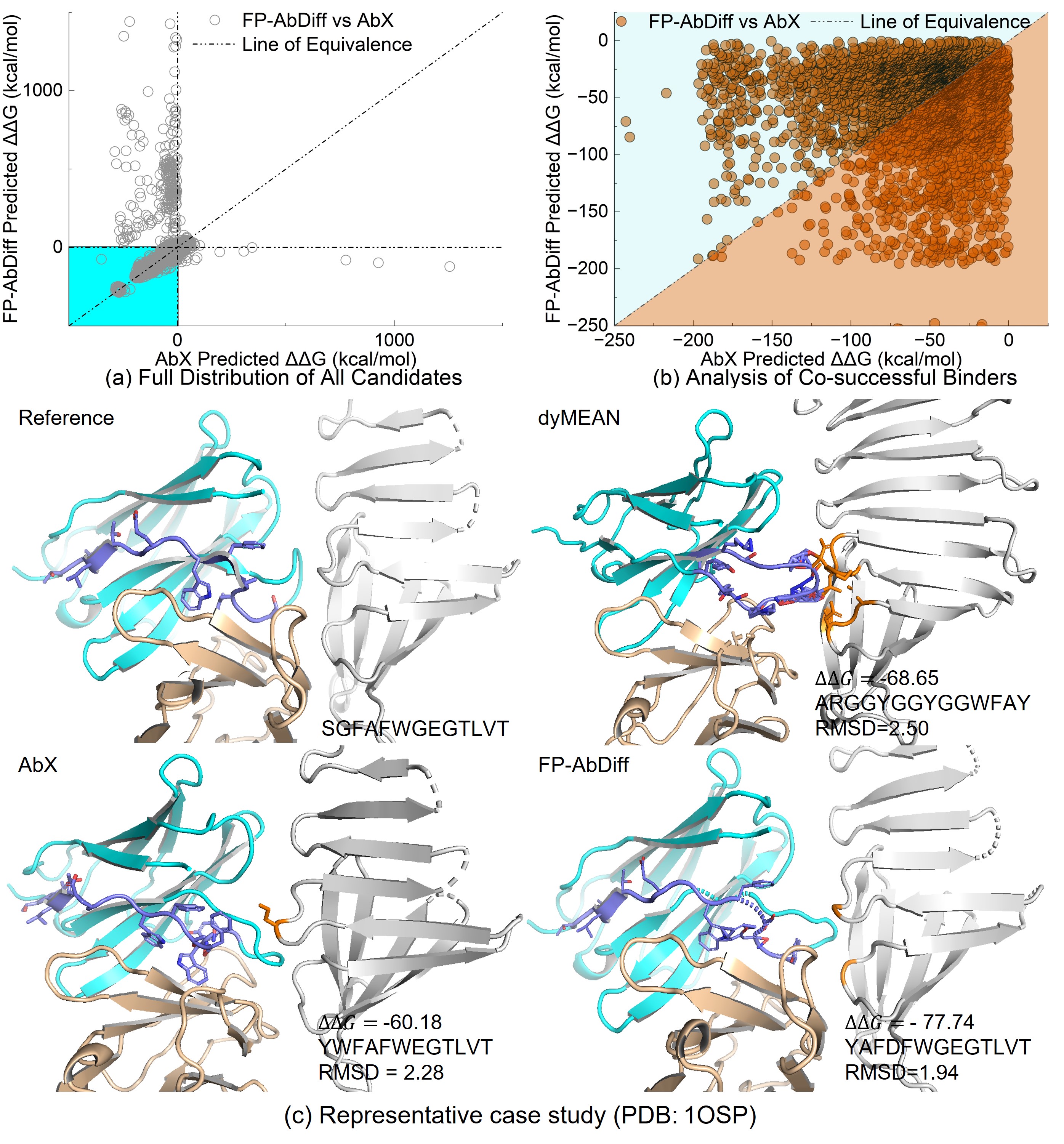}
    \caption{
    Binding affinity comparison and representative structure design.
    \textbf{(a)} Predicted binding energy change ($\Delta\Delta G$) for all designs after full-CDR relaxation. Cyan quadrant: designs with $\Delta\Delta G<0$ for both FP-AbDiff and AbX.
    \textbf{(b)} Zoomed view of (a), highlighting co-successful cases.
    \textbf{(c)} Representative example on PDB: 1OSP.
    }
    \label{fig:affinity_and_case_study}
\end{figure}

\paragraph{Energetic Validation and Case Study.}
To assess energetic performance, we compared the predicted binding affinities ($\Delta\Delta G$) of FP-AbDiff and the AbX baseline across thousands of designs (Figure~\ref{fig:affinity_and_case_study} (a) and (b)). The roughly symmetric distribution around the line of equivalence indicates comparable energy prediction capabilities. Notably, FP-AbDiff retains AbX-level performance despite its physics-driven regularization, validating that physical consistency does not compromise functional fidelity. The advantage becomes clearer when considering structure and energy jointly: in a representative case (Figure~\ref{fig:affinity_and_case_study} (c)), FP-AbDiff achieves lower RMSD and favorable interface energy, demonstrating co-optimization of geometry and binding without trade-offs.

\subsection{Antibody Optimization}
\label{sec:affinity}

To assess long-horizon optimization, iterative denoising was applied to perturbed CDR-H3s from the RAbD test set, targeting improved binding affinity ($\Delta\Delta G$) while maintaining structural plausibility. As shown in Table~\ref{tab:optimization_trajectory}, a key trade-off emerges between energy gain and structural stability. AbX follows a “high-gain but brittle” trajectory—peaking early ($t \leq 16$) in IMP but degrading from 1.13 \AA{} to 2.88 \AA{} in RMSD. In contrast, FP-AbDiff maintains consistently lower RMSD and higher DockQ from $t = 8$ onward, reflecting a more stable optimization path. This resilience (enabled by Fokker–Planck regularization) ensures affinity gains arise from structurally viable conformations, promoting safe and experimentally tractable designs.

\begin{table}[t]
\centering

\small
\setlength{\tabcolsep}{3.3pt} 

\begin{tabular}{llcccc}
\hline
\textbf{Model} & \textbf{Steps} & \textbf{IMP(\%)↑} & \textbf{AAR(\%)↑} & \textbf{RMSD(\AA)↓} & \textbf{DockQ↑} \\
\hline
\multirow{7}{*}{\begin{tabular}{@{}c@{}}\textbf{FP-}\\\textbf{AbDiff}\end{tabular}} 
& 2  & 36.04 & 43.07 & 1.23 & 0.4809 \\
& 4  & 35.79 & 43.27 & 1.29 & 0.4789 \\
& 8  & 32.71 & 43.42 & 1.41 & 0.4639 \\
& 16 & 25.06 & 44.17 & 2.09 & 0.4265 \\
& 32 & 26.67 & 44.25 & 2.23 & 0.4330 \\
& 64 & 28.62 & 44.27 & 2.25 & 0.4416 \\
& T  & 28.42 & 44.34 & 2.24 & 0.4443 \\
\hline
\multirow{7}{*}{\textbf{AbX}} 
& 2  & 47.38 & 45.87 & 1.13 & 0.4981 \\
& 4  & 45.90 & 46.40 & 1.29 & 0.4870 \\
& 8  & 46.19 & 46.23 & 1.48 & 0.4765 \\
& 16 & 46.39 & 45.67 & 1.75 & 0.4652 \\
& 32 & 44.04 & 45.15 & 2.16 & 0.4512 \\
& 64 & 41.75 & 44.19 & 2.68 & 0.4341 \\
& T  & 41.38 & 44.00 & 2.88 & 0.4287 \\
\hline
\end{tabular}
\caption{
    Performance on the CDR-H3 optimization task. All metrics are calculated on the full chain ($^{Full}$).
}
\label{tab:optimization_trajectory}
\end{table}

\subsection{Ablation Studies}
\label{sec:ablation}

\begin{table}[htbp!]
\centering
\small
\setlength{\tabcolsep}{3.1pt}

\begin{tabular}{lcccc}
\hline
\textbf{Model Variant} & \textbf{IMP(\%)↑} & \textbf{AAR(\%)↑} & \textbf{RMSD(\AA)↓} & \textbf{DockQ↑} \\
\hline
+R3, +SO(3)   & 28.42          & \textbf{45.23} & \textbf{2.18} & \textbf{0.4443} \\
- SO(3)        & \textbf{35.30} & 44.15          & 2.46          & 0.4437 \\
- R3           & \underline{29.76}    & \underline{43.14}    & \underline{2.41}    & \underline{0.4372} \\
\hline
\end{tabular}
\caption{
    Ablation study on the RAbD CDR-H3 co-design task.
    All metrics reported are for the CDR-H3 loop and are calculated on the full chain ($^{Full}$).
}
\label{tab:ablation_study}
\end{table}
To dissect the role of Fokker–Planck regularization on translational (\(\mathbb{R}^3\)) and rotational (\(\mathrm{SO}(3)\)) manifolds, ablations are performed on the RAbD CDR-H3 co-design task (Table~\ref{tab:ablation_study}). The full model yields the highest fidelity, with lowest RMSD and highest DockQ. Removing the \(\mathbb{R}^3\) term degrades both backbone and interface quality, highlighting its role in constraining atomic positions. In contrast, dropping the \(\mathrm{SO}(3)\) term paradoxically increases IMP to 35.30\% despite worse RMSD and AAR. The apparent gain reflects refinability, not fidelity: the –SO(3) variant outputs strained, high-energy poses that relax into lower energies and inflate IMP, whereas the full model emits physically sound structures from the outset, leaving little room for such artificial gains.

\section{Conclusion}
\label{sec:conclusion}

We introduced FP-AbDiff, a Fokker–Planck-regularized antibody diffusion model that enforces globally consistent and physically plausible dynamics within an SE(3)-equivariant framework. Our method consistently outperforms state-of-the-art baselines across all evaluated design tasks, delivering high-fidelity structures, precise interfaces, and stable generative trajectories. Ablation studies confirm that these comprehensive performance gains are a direct consequence of imposing dynamical consistency on the $\mathbb{R}^3 \times \mathrm{SO}(3)$ manifold. This enables an approach shift towards correct-by-construction generation, reducing dependence on post-hoc refinement and establishing a new benchmark for physically grounded antibody design. While awaiting experimental validation and offering opportunities for refining its numerical approximations, FP-AbDiff's synergistic integration of physical laws with deep biological priors establishes a robust and generalizable foundation for the next generation of truly physically faithful immuno-engineering.

\bibliography{arxiv}


\end{document}